\documentclass{article}

\usepackage[final]{nips_2016}

\usepackage[utf8]{inputenc} 
\usepackage[T1]{fontenc}    
\usepackage{hyperref}       
\usepackage{url}            
\usepackage{booktabs}       
\usepackage{amsfonts}       
\usepackage{nicefrac}       
\usepackage{microtype}      
\usepackage{float}
\usepackage{caption}
\usepackage{subcaption}
\usepackage{graphicx}
\usepackage{amsmath}
\usepackage[toc,page]{appendix}
\usepackage{color}

\title{CycleGAN Face-off}

\author{
  Xiaohan Jin* \\
  \texttt{xiaohanj@andrew.cmu.edu} \\
  \And
  Ye Qi* \\
  \texttt{yeq@andrew.cmu.edu} \\
  \And 
  Shangxuan Wu* \\
  \texttt{shangxuw@andrew.cmu.edu} \\
}

\begin{document}

\maketitle

\section{Introduction}
Face-off is an interesting case of style transfer where the facial expressions and attributes of one person could be fully transformed to another face. We are interested in the unsupervised training process which only requires two sequences of unaligned video frames from each person and learns what shared attributes to extract automatically. In this project, we explored various improvements for adversarial training (i.e. CycleGAN\citep{zhu2017unpaired}) to deal with the common problem of model collapse,  to capture details in facial expressions and head poses, and thus transfer facial expressions with higher consistency and stability.
\begin{figure}[H]
\centering
\begin{subfigure}{0.5\textwidth}
  \centering
   \includegraphics[width=\linewidth]{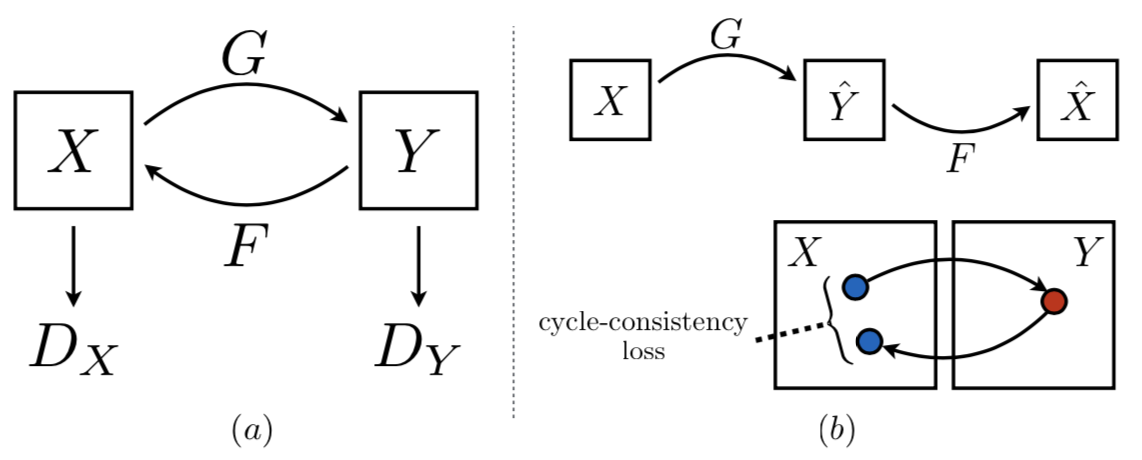}
  \caption*{Vanilla CycleGAN\citep{zhu2017unpaired} structure.}
  \label{fig:sub1}
\end{subfigure}%
\begin{subfigure}{0.5\textwidth}
  \centering
  \includegraphics[width=.55\linewidth]{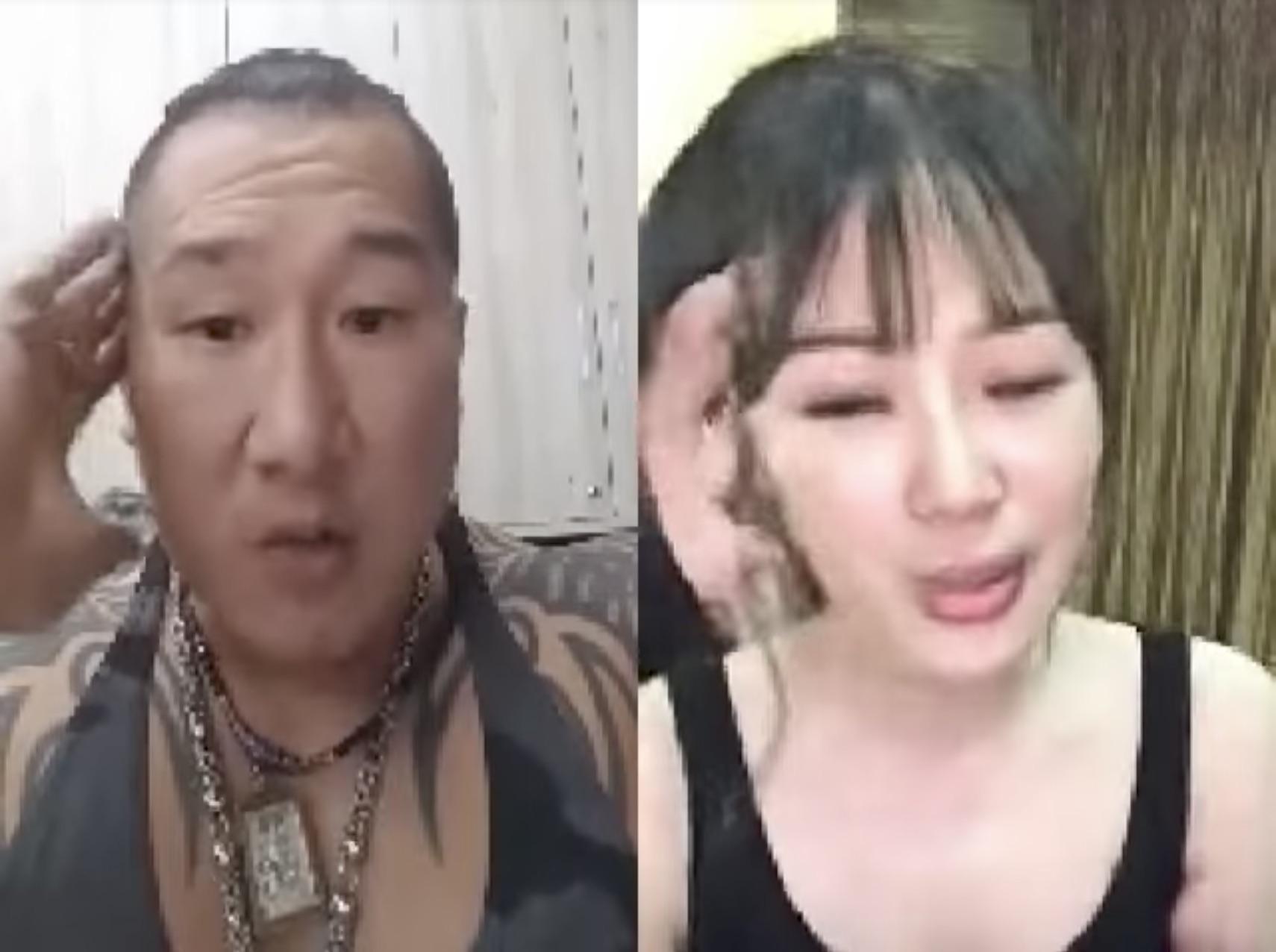}
  \caption*{CycleGAN face-off application ~\citep{tjwei2017youtube}.}
  \label{fig:sub2}
\end{subfigure}
\label{fig:test}
\end{figure}
\subsection{Our Contribution}
This project\footnote{\textcolor{blue}{Please see this video first (use headphone!): \\ \url{https://drive.google.com/file/d/1QFh-n0L6Q-3kwfO1WIKjf7_O5dsCwClL/view?usp=sharing}}}\footnote{Code: \url{https://github.com/ShangxuanWu/CycleGAN-Face-off}} explored different approaches to generate better face-off videos based on CycleGAN\citep{zhu2017unpaired}. Our experiment results are summarized as below:
\begin{enumerate}
    \item We proposed several working methods for improving CycleGAN face-off results:
    \begin{enumerate}
    \item Using two distinct discriminators\citep{xu2017face}.
    \item Using face segmentation masks as weight on cycle-consistency loss.
    \end{enumerate}
    \item We also concluded that following variants might not improve face-off results on our current dataset:
    \begin{enumerate}
    \item Applying WGAN loss\citep{gulrajani2017improved} to tweak the network.
    \item Using U-Net\citep{ronneberger2015u} as generator.
    \end{enumerate}
    \item Using SSIM loss\citep{wang2003multiscale} with loss weight within a proper range learns better poses and structure, but needs more tuning to improve facial details.
    
\end{enumerate}
\section{Related Work}
Facial expression transfer is a classic topic in computer vision and graphics using facial landmark localization and face morphing, and it became more popular with the adaptation of neural style transfer and GANs ~\citep{zhu2017unpaired}~\citep{goodfellow2014generative}~\citep{radford2015unsupervised}~\citep{isola2016image}. GAN models such as Pix2pix~\citep{isola2016image} and CycleGAN~\citep{zhu2017unpaired} have shown impressive results on image-to-image translation that learns to relate two different data domains. 

\subsection{CycleGAN}
The loss function of CycleGAN is composed of two parts: traditional GAN loss and a new cycle-consistency loss which pushes cycle consistency:
\begin{align}
L(G,F, D_X, D_Y) =\  &L_{GAN}(G, D_Y, X, Y)\ +\notag  \\ 
					&L_{GAN}(F, D_X, Y, X)\ + \notag \\
                    &\lambda L_{cyc}(G,F)
\end{align}
where cycle-consistency loss represents how similar the $G(F(X))$ is to $X$ and $F(G(Y))$ is to $Y$:
\begin{equation}
L_{cyc}(G,F) = E_{x~p_{data}(x)}[||F(G(x))-x||_1]+E_{y~p_{data}(y)}[||G(F(y))-y||_1]
\end{equation}

However in our experiments, we have found it very challenging to get good transferring results on unaligned datasets. So in the next section, we will discuss some different techniques to deal with it.

\section{Methods}
In this section, we explore how to generate better face-off sequences on unaligned datasets in the following aspects: (1) using WGAN loss~\citep{gulrajani2017improved} in adversarial part; (2) adding SSIM loss~\citep{wang2003multiscale} in cycle-consistency loss; (3) adding face segmentation masks as reconstruction weights; (4) using skip layers to increase multi-scale invariance; (5) increasing the capacity of discriminators, and etc.

\subsection{WGAN Loss}
We consider WGAN ~\citep{gulrajani2017improved} to deal with the common problem of model collapse in adversarial training and to achieve more stable results. As we found from our tests, some of the expressions of person A were transferred to the same pose and expression of person B. The standard discriminator loss uses cross-entropy loss and suffers from gradient vanishing. To solve this, we adopted the following improvements according to WGAN paper:
\begin{itemize}
\itemsep0em
\item No $log$ in the loss. The output of $D$ is no longer a probability, hence we do not apply sigmoid at the output of $D$.
\item Clip the weight in $D$.
\item Train $D$ more than $G$.
\item Use RMSProp instead of ADAM.
\item Lower learning rate. The rate in the paper is $\alpha = 0.00005$.
\end{itemize}

\subsection{SSIM (Structural Similarity) Loss}

SSIM loss~\citep{wang2003multiscale} matches the luminance($l$), contrast($c$), and structure($s$) information of the generated image and the input image, and it's proved to be very helpful to improve the quality of image generation. Multi-Scale SSIM loss considers SSIM loss over M scales as follows:
\begin{equation}
MS-SSIM(x,y) = [l_M(x,y)]^{\alpha_M}\dot \prod_{j=1}^M[c_j(x,y)]^{\beta_j}[s_j(x,y)]^{\gamma_j} 
\end{equation}
where
\begin{align}
    l(x,y)&=\frac{2\mu_x\mu_y + c_1}{\mu_x^2+\mu_y^2 + c_1}\\
    c(x,y)&=\frac{2\sigma_x\sigma_y + c_1}{\sigma_x^2+\sigma_y^2 + c_1}\\
    s(x,y)&=\frac{\sigma_{xy} + c_3}{\sigma_x\sigma_y+ c_3}
\end{align}

\begin{figure}[H]
\centering
\includegraphics[width=0.8\linewidth]{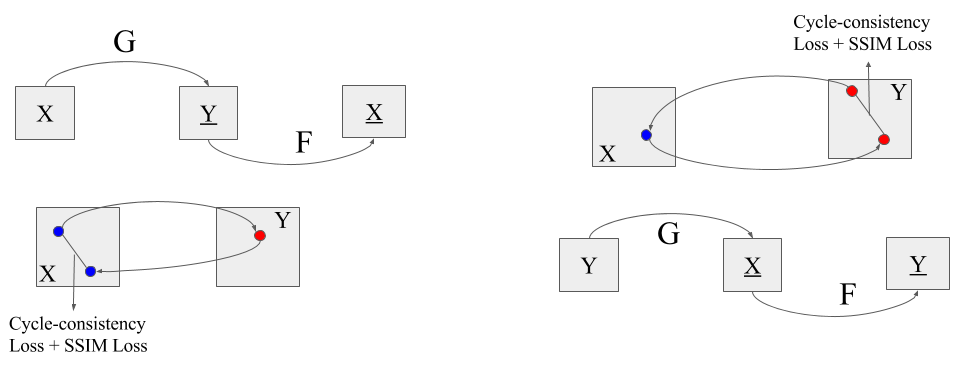}
\caption{Structural similarity loss adding to CycleGAN structure.}
\label{fig:ssim}
\end{figure}

We added SSIM loss in CycleGAN in order to enforce the similarity between recovered image and original image (as illustrated in Figure \ref{fig:ssim}). 

\subsection{Background Subtraction and Face Mask}

We observed severe background corruption in the video shown in ~\citep{tjwei2017youtube}.  One possible reason is that rather than explicitly separating foreground and background, CycleGAN treats the whole image as one object and transfers the domain implicitly.

\begin{figure}[h!]
\centering
\includegraphics[width=0.8\linewidth]{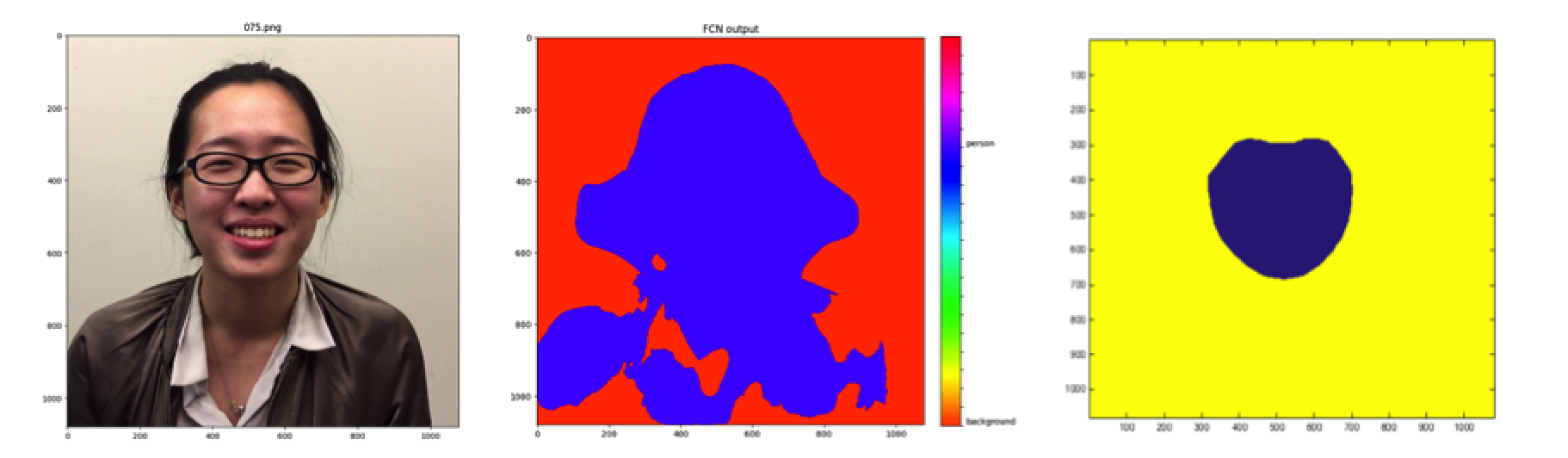}
\caption{Illustration of segmentation masks, from left to right: input, FCN person segmentation, face mask generated from DLib facial landmarks.}
\label{fig:seg}
\end{figure}

Therefore, in this project, we explicitly dealt with foreground and background, trying to get a sharper boundary of the target. We segmented the input faces as shown in Figure~\ref{fig:seg}, then fed the mask as weight on pixel-wise reconstruction error. 

We experimented with both fully convolutional networks (FCN)\citep{long2015fully} and DLib\citep{dlib} in this project. Using FCN-8s pretrained on Pascal VOC dataset, we can segment the whole person from the image, which guides the network to put more focus on the whole body. However, we noticed that the segmentation result is very unstable, and cannot give a very clear contour of a person. So we later looked at DLib by which we extracted the facial landmarks and then transformed the face polygons to masks. Using the face masks are more helpful for focusing on facial expressions.

There are two instinctive approaches to leverage face mask:
\begin{enumerate}
\item Crop out face part only for network input and neglect all the other parts.
\item Based on segmentation mask, apply pixel-wise weight to original cycle-consistency loss.
\end{enumerate}

We experimented with both of them and results will be analyzed in the next session.

\subsection{Generator Variants}
\begin{figure}[H]
	\centering
      \begin{minipage}{0.55\linewidth}
          \begin{figure}[H]
              \includegraphics[width=\linewidth]{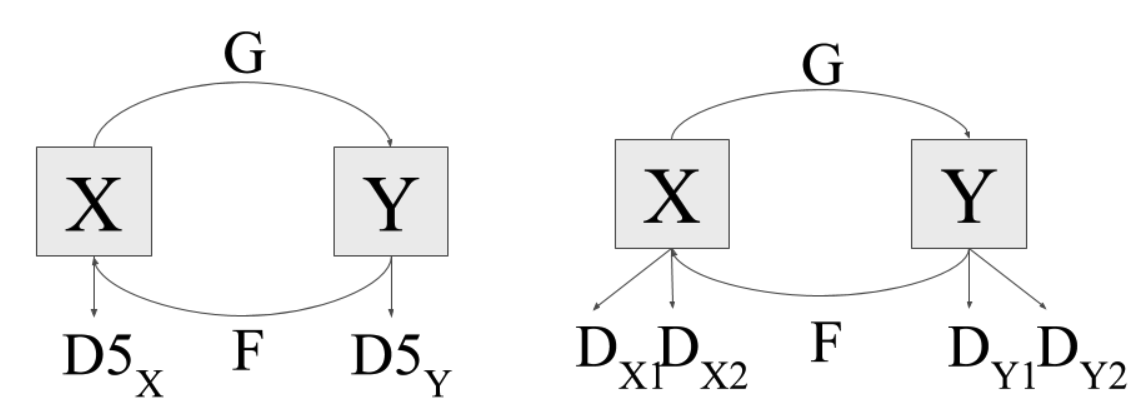}
          \end{figure}
      \end{minipage}
      \hspace{0.05\linewidth}
      \begin{minipage}{0.35\linewidth}
          \begin{figure}[H]
              \includegraphics[width=\linewidth]{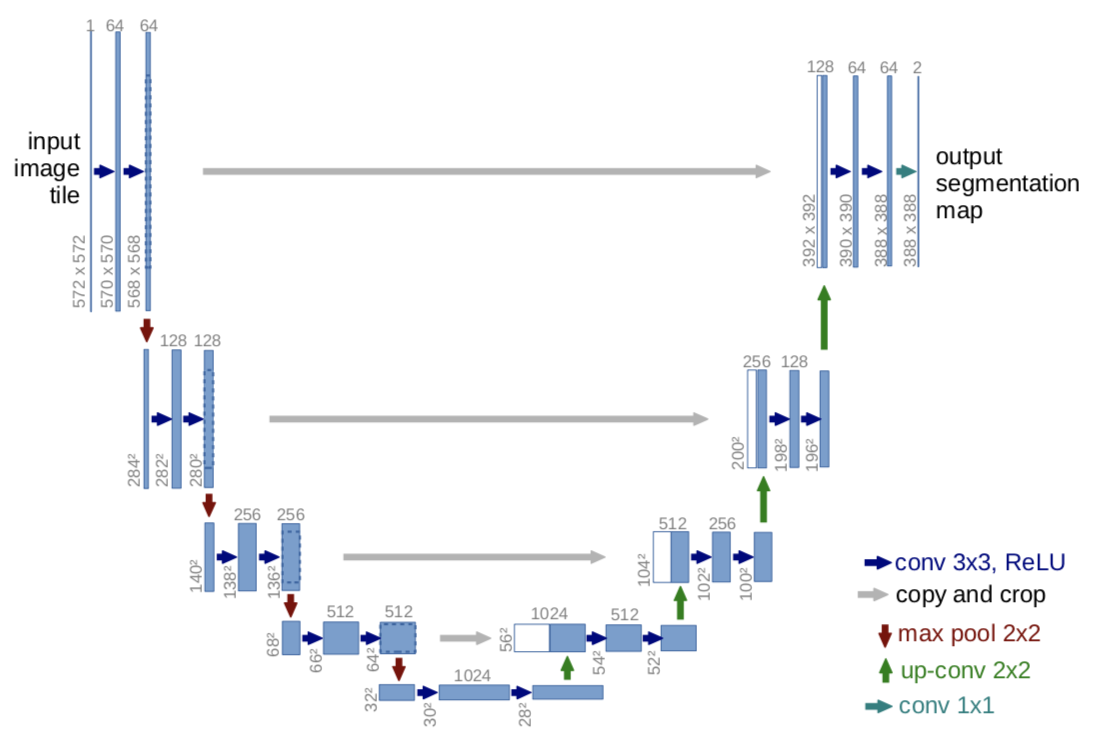}
          \end{figure}
      \end{minipage}
      \caption{\textbf{Left two} Discriminator variants: deeper discriminators and multiple discriminators. \textbf{Right} Generator variants: U-Net \citep{ronneberger2015u} structure.}
      \label{Variants}
\end{figure}

We want to explore different generator structures. A widely adopted variant would be U-Net\citep{ronneberger2015u} with skip layers. The structure of U-Net is shown in Figure \ref{Variants}.

\subsection{Discriminator Variants}
According to \citep{xu2017face}, improving the capacity of discriminator in GANs would result in more natural and higher resolution image generation. Therefore, we propose some ways to enforce the discriminative process:
\begin{enumerate}
\item Vanilla CycleGAN uses 3-layer-Conv $D_X$ and $D_Y$ as discriminator. We will first to extend the depth of this subnetwork to 5-layer-Conv.
\item Alternatively, we can use two different discriminators at each side, and average their loss using a given weight $\lambda$. $\lambda$ is set as 0.5 in our experiments. The loss function is modified as:
\end{enumerate}
\begin{align}
L_{GAN}(G, D_{Y1}, D_{Y2}) = \lambda L_{GAN}(G, D_{Y1}) + (1 - \lambda) L_{GAN}(G, D_{Y2})
\end{align}



\section{Dataset}
We evaluated proposed methods on four videos: 
\begin{itemize}
\item[A] A 90-second clip cropped from Prof. Russ Salakhutdinov's talk on deep learning\footnote{\url{https://www.youtube.com/watch?v=2iMemrQm-io}}.
\item[B] A video of Shangxuan talking to the camera shot by ourselves. 
\item[C] A video of Ye talking to the camera shot by ourselves. 
\item[D] A video of Xiaohan talking to the camera shot by ourselves. 
\end{itemize}

\begin{figure}[H]
\centering
\begin{subfigure}{0.12\textwidth}
  \centering
  \includegraphics[width=\linewidth]{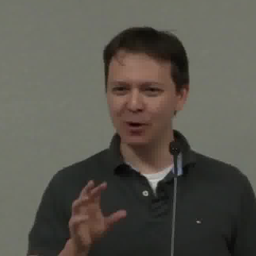}
\end{subfigure}\hfill
\begin{subfigure}{0.12\textwidth}
  \centering
  \includegraphics[width=\linewidth]{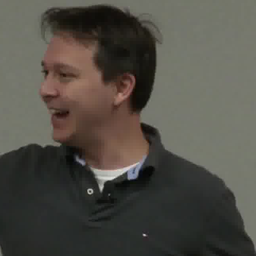}
\end{subfigure}\hfill
\begin{subfigure}{0.12\textwidth}
  \centering
  \includegraphics[width=\linewidth]{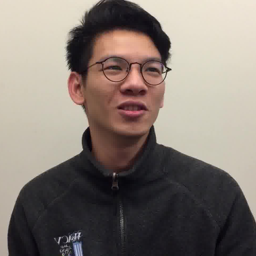}
\end{subfigure}\hfill
\begin{subfigure}{0.12\textwidth}
  \centering
  \includegraphics[width=\linewidth]{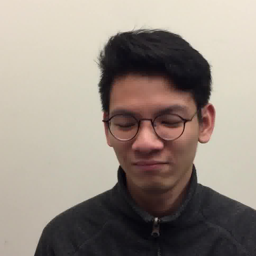}
\end{subfigure}\hfill
\begin{subfigure}{0.12\textwidth}
  \centering
  \includegraphics[width=\linewidth]{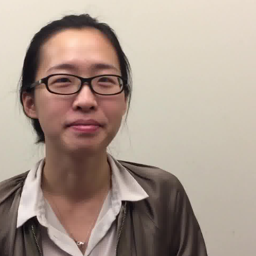}
\end{subfigure}\hfill
\begin{subfigure}{0.12\textwidth}
  \centering
  \includegraphics[width=\linewidth]{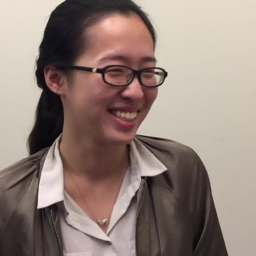}
\end{subfigure}\hfill
\begin{subfigure}{0.12\textwidth}
  \centering
  \includegraphics[width=\linewidth]{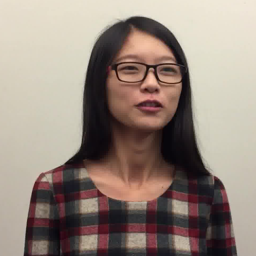}
\end{subfigure}\hfill
\begin{subfigure}{0.12\textwidth}
  \centering
  \includegraphics[width=\linewidth]{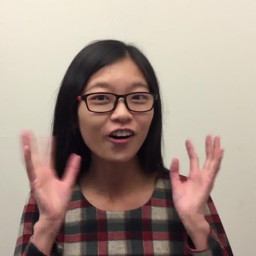}
\end{subfigure}
\caption{Example of 4 datasets. During training, the first two images are used as input A; the other 3 datasets are used as input B.  }
\label{fig:dataset}
\end{figure}

We created four datasets\footnote{\url{https://drive.google.com/drive/folders/1Oepq5qBleF9mDrzdulEzWnY04FAlPUA1?usp=sharing}}\footnote{\url{https://drive.google.com/drive/folders/1yGZ0NJJeqxhSYONIUjOc7v30NhpAS65U?usp=sharing}} from the videos described above by extracting 2,200 frames from each clip, in which 2,000 of them belong to the training set while 200 are testing images. All videos were shot against relatively plain background but the faces in each are not perfectly aligned, as shown in Figure~\ref{fig:dataset}. Each image is of resolution $286 \times 286$, and will be randomly cropped to $256 \times 256$ when being input to the network for data augmentation. Our face-off experiments were performed on three pairs of datasets A2B, A2C, A2D. 

\section{Experiments and Result Analysis}
\begin{figure}
\centering
\begin{subfigure}{0.18\textwidth}
  \centering
  \includegraphics[width=\linewidth]{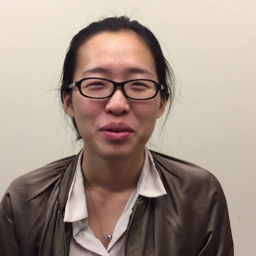}
  \caption*{Input}
\end{subfigure}\hfill
\begin{subfigure}{0.18\textwidth}
  \centering
  \includegraphics[width=\linewidth]{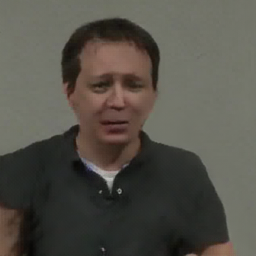}
  \caption*{Baseline}
\end{subfigure}\hfill
\begin{subfigure}{0.18\textwidth}
  \centering
  \includegraphics[width=\linewidth]{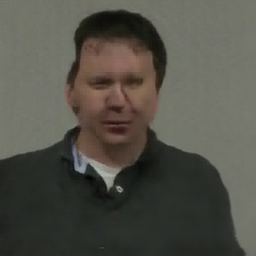}
  \caption*{With SSIM}
\end{subfigure}\hfill
\begin{subfigure}{0.18\textwidth}
  \centering
  \includegraphics[width=\linewidth]{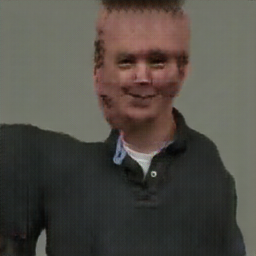}
  \caption*{U-Net as G}
\end{subfigure}\hfill
\begin{subfigure}{0.18\textwidth}
  \centering
  \includegraphics[width=\linewidth]{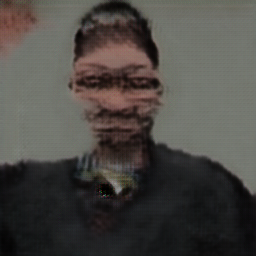}
  \caption*{WGAN}
\end{subfigure}
\vfill
\begin{subfigure}{0.18\textwidth}
  \centering
  \includegraphics[width=\linewidth]{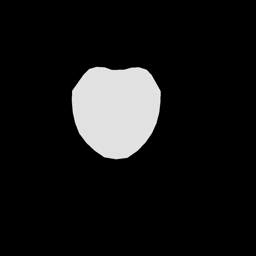}
  \caption*{Face Seg Mask}
\end{subfigure}\hfill
\begin{subfigure}{0.18\textwidth}
  \centering
  \includegraphics[width=\linewidth]{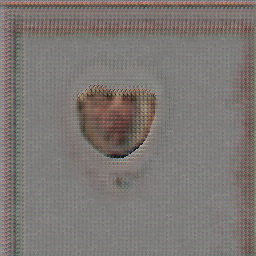}
  \caption*{Crop Mask}
\end{subfigure}\hfill
\begin{subfigure}{0.18\textwidth}
  \centering
  \includegraphics[width=\linewidth]{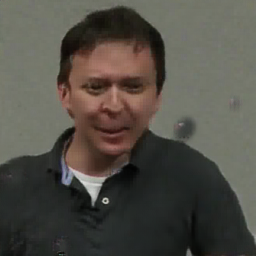}
  \caption*{Weight Mask}
  \label{fig:sub1}
\end{subfigure}\hfill
\begin{subfigure}{0.18\textwidth}
  \centering
  \includegraphics[width=\linewidth]{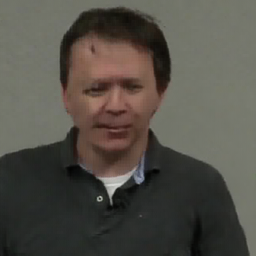}
  \caption*{5-layer D}
  \label{fig:sub2}
\end{subfigure}\hfill
\begin{subfigure}{0.18\textwidth}
  \centering
  \includegraphics[width=\linewidth]{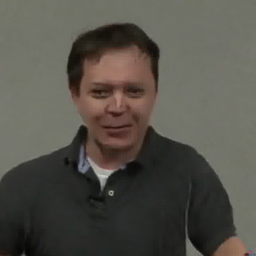}
  \caption*{Double D (3+5)}
  \label{fig:sub1}
\end{subfigure}
\vfill\vfill\vfill
\begin{subfigure}{0.18\textwidth}
  \centering
  \includegraphics[width=\linewidth]{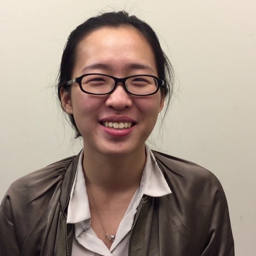}
  \caption*{Input}
\end{subfigure}\hfill
\begin{subfigure}{0.18\textwidth}
  \centering
  \includegraphics[width=\linewidth]{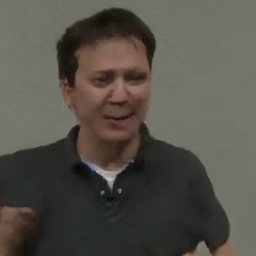}
  \caption*{Baseline}
\end{subfigure}\hfill
\begin{subfigure}{0.18\textwidth}
  \centering
  \includegraphics[width=\linewidth]{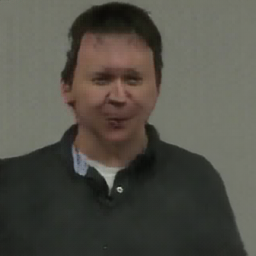}
  \caption*{With SSIM}
\end{subfigure}\hfill
\begin{subfigure}{0.18\textwidth}
  \centering
  \includegraphics[width=\linewidth]{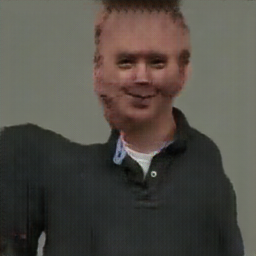}
  \caption*{U-Net as G}
\end{subfigure}\hfill
\begin{subfigure}{0.18\textwidth}
  \centering
  \includegraphics[width=\linewidth]{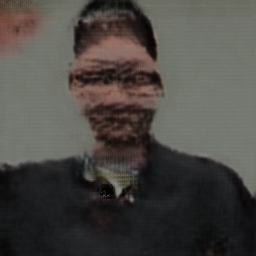}
  \caption*{WGAN}
\end{subfigure}
\vfill
\begin{subfigure}{0.18\textwidth}
  \centering
  \includegraphics[width=\linewidth]{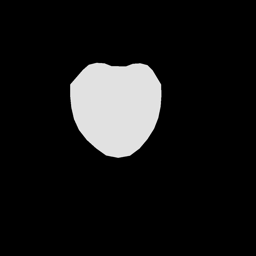}
  \caption*{Face Seg Mask}
\end{subfigure}\hfill
\begin{subfigure}{0.18\textwidth}
  \centering
  \includegraphics[width=\linewidth]{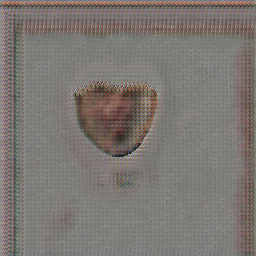}
  \caption*{Crop Mask}
\end{subfigure}\hfill
\begin{subfigure}{0.18\textwidth}
  \centering
  \includegraphics[width=\linewidth]{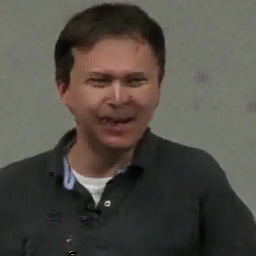}
  \caption*{Weight Mask}
  \label{fig:sub1}
\end{subfigure}\hfill
\begin{subfigure}{0.18\textwidth}
  \centering
  \includegraphics[width=\linewidth]{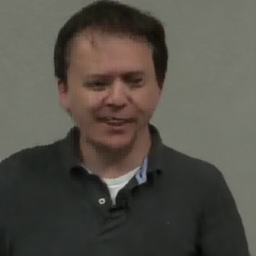}
  \caption*{5-layer D}
  \label{fig:sub2}
\end{subfigure}\hfill
\begin{subfigure}{0.18\textwidth}
  \centering
  \includegraphics[width=\linewidth]{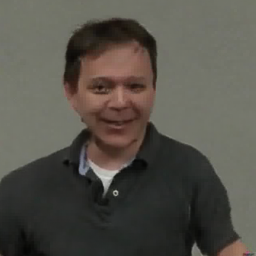}
  \caption*{Double D (3+5)}
  \label{fig:sub1}
\end{subfigure}
\vfill\vfill\vfill
\begin{subfigure}{0.18\textwidth}
  \centering
  \includegraphics[width=\linewidth]{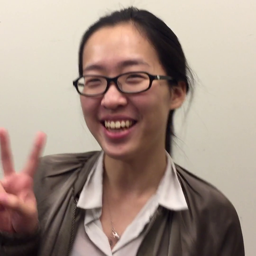}
  \caption*{Input}
\end{subfigure}\hfill
\begin{subfigure}{0.18\textwidth}
  \centering
  \includegraphics[width=\linewidth]{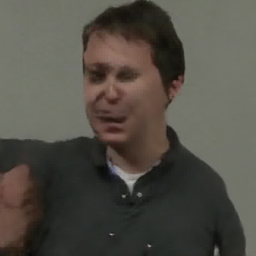}
  \caption*{Baseline}
\end{subfigure}\hfill
\begin{subfigure}{0.18\textwidth}
  \centering
  \includegraphics[width=\linewidth]{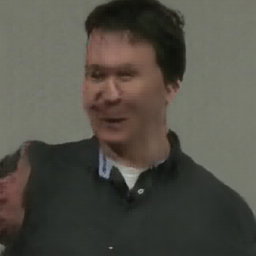}
  \caption*{With SSIM}
\end{subfigure}\hfill
\begin{subfigure}{0.18\textwidth}
  \centering
  \includegraphics[width=\linewidth]{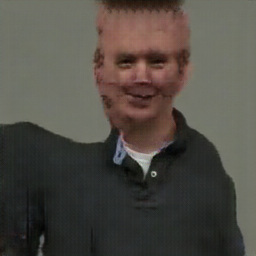}
  \caption*{U-Net as G}
\end{subfigure}\hfill
\begin{subfigure}{0.18\textwidth}
  \centering
  \includegraphics[width=\linewidth]{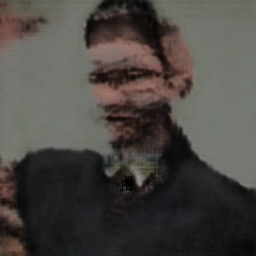}
  \caption*{WGAN}
\end{subfigure}
\vfill
\begin{subfigure}{0.18\textwidth}
  \centering
  \includegraphics[width=\linewidth]{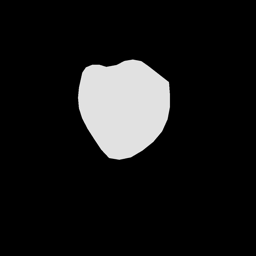}
  \caption*{Face Seg Mask}
\end{subfigure}\hfill
\begin{subfigure}{0.18\textwidth}
  \centering
  \includegraphics[width=\linewidth]{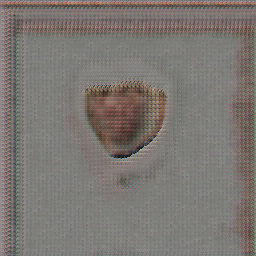}
  \caption*{Crop Mask}
\end{subfigure}\hfill
\begin{subfigure}{0.18\textwidth}
  \centering
  \includegraphics[width=\linewidth]{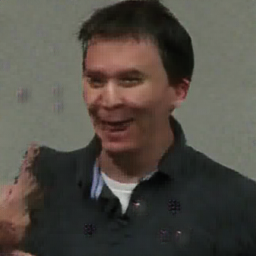}
  \caption*{Weight Mask}
  \label{fig:sub1}
\end{subfigure}\hfill
\begin{subfigure}{0.18\textwidth}
  \centering
  \includegraphics[width=\linewidth]{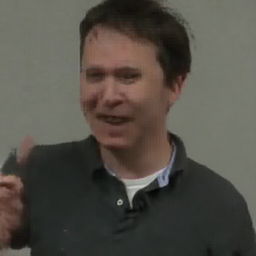}
  \caption*{5-layer D}
  \label{fig:sub2}
\end{subfigure}\hfill
\begin{subfigure}{0.18\textwidth}
  \centering
  \includegraphics[width=\linewidth]{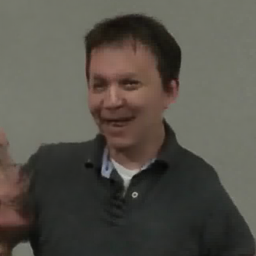}
  \caption*{Double D (3+5)}
  \label{fig:sub1}
\end{subfigure}
\caption{CycleGAN testing results on multiple settings.}
\label{fig:test_results}
\end{figure}

The results of different experiments on baseline, SSIM loss, WGAN loss, UNet structure, face masks and discriminator structures are shown in Figure~\ref{fig:test_results}, of which analysis is as follows.

\subsection{Baseline}
We trained a baseline CycleGAN model with 3-layer discriminator (3 convolution blocks). We back-propagates using least-square GAN (LS-GAN) loss and cycle consistency ($L_1$) loss. This was trained on unaligned dataset where two people appear with different poses and at different spatial locations as mentioned in the Dataset section. 

rFor face-off between Shangxuan and Russ, the baseline model already learns the mapping of human poses well, but the results are quite unstable at edges. Since the faces of Shangxuan and Russ are not perfectly aligned, the model does not deal with scale well.

For face-off between Ye and Russ, the baseline model learns to transform well between different gender domains. But the results are  very shaky and inconsistent between the frames as well. 

For face-off between Xiaohan and Russ, the baseline model learns to transform human poses and gender as well, but maps Xiaohan's hairstyle to Russ, thus creates a lot of noisy results.

\subsection{WGAN Loss}
We implemented WGAN loss to improve the training of GANs. However, the training was very unstable, even if we tuned the learning rate and clipped the gradients. Using WGAN has a high failure rate and is slow to train. We can see the results that least-square GAN loss helps produce better results compared to WGAN loss. 

\subsection{SSIM Loss}
SSIM loss should have a better interpretation of matching the luminance, contrast, and structure information. Though it has been shown to perform perceptually preferable images, it is not shown to improve the image details in our experiments. After some tuning, we found a proper range for weights of SSIM should be around 0.0001 to 0.01. The weight should not be too large, or it might dominate the reconstruction loss and we cannot see as much details as the baseline model using $L_1$ loss. We added SSIM loss with weight 0.01, we can see in Figure~\ref{fig:test_results} that using SSIM can help learn the pose well, while it still needs more tuning to recover more facial details, but generally it is helpful.

\subsection{Face Mask}
We want the model to learn to focus on facial expressions, and allow higher gradient flow to train those areas better. So we experimented with face masks generated by DLib. There are two ways to use the face masks. First is to crop the input images and input the masked image; second is to apply the mask to the loss function.

\subsubsection{Mask-out Background}
When we mask out the background, the face left is only a small portion in the image that the model has to learn. In such a scenario the generator exhibits very poor diversity amongst generated samples so the discriminator cannot tag them as fake, which makes training discriminator ineffective. The model got collapsed easily. Common ways to solve this problem is to increase the dataset diversity or batch size, to let the discriminator learn more about edge-cases, or use WGAN to improve the training of GANs. However, we tried using WGAN but that is hard to tune as well. 

\subsubsection{Mask as Loss Weight}
From the previous experiment, we think that it is better to keep the background but more importantly, to increase the weight of facial parts. Thus during training, we input binary masks together with images and applied element wise product of $w_{mask}I_{mask}+1$ with $L_1$ reconstruction loss. We can see from Figure~\ref{fig:test_results} that, this can add more facial details such as teeth as well as more natural facial expressions. With higher gradient flow on faces, the network learns to focus on facial details more.

\subsection{U-Net in Generator}

We observed a severe mode collapse using U-Net as generator. The generated images are nearly identical even if the input images have different poses. The reason is that comparing to vanilla ResNet generator structure, U-Net is not so capable of extracting image representations. We would suggest using a stronger network if we are to substitute the generator in the future.

\subsection{Multiple Deeper Discriminators}

When the number of discriminator layer increases, the receptive field size is reduced, forcing the model to learn a more detailed translation from one domain to another. The result demonstrates that the model leveraging 5-layer discriminator does a better job at imitating the facial expressions of the input though the global structure like head-shoulder ratio is not held as well. 

Multiple-discriminator GAN, as appose to the one with a single discriminator, amplifies the model capacity and resists random noises. It noticeably outperforms other settings when encountering an image with an unseen pose (as revealed by the fifth and sixth row in Figure~\ref{fig:test_results}). With a reasonable trade-off between patterns learned from different receptive fields, the generator perfectly combines the subtle facial expressions from the source person without detouring too much from target person's features. 

\subsection{Training Loss Comparison}

We plotted the three training losses (generator loss, discriminator loss and cycle-consistency loss) for three different architectures: baseline (vanilla CycleGAN), a good face-off network (Double Discriminator CycleGAN) and a bad face-off network (U-Net Generator CycleGAN). The plots are shown below:

\begin{figure}[H]
\centering
\begin{subfigure}{0.32\textwidth}
  \centering
  \includegraphics[width=\linewidth]{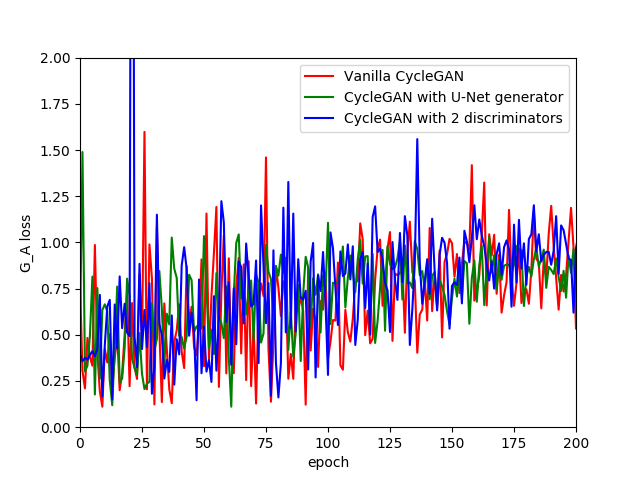}
  \caption*{Generator Loss}
  \label{fig:sub1}
\end{subfigure}\hfill
\begin{subfigure}{0.32\textwidth}
  \centering
  \includegraphics[width=\linewidth]{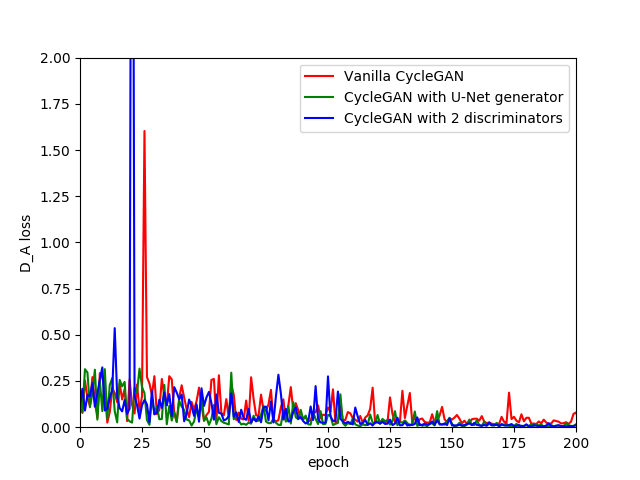}
  \caption*{Discriminator Loss}
  \label{fig:sub2}
\end{subfigure}\hfill
\begin{subfigure}{0.32\textwidth}
  \centering
  \includegraphics[width=\linewidth]{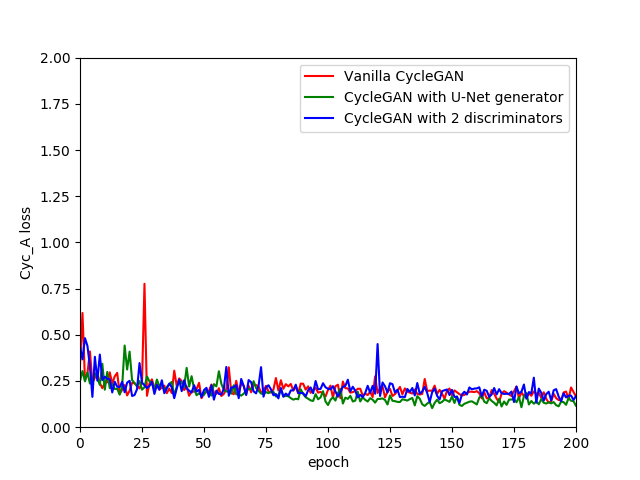}
  \caption*{Cycle-consistency Loss}
  \label{fig:sub1}
\end{subfigure}\hfill
\caption{CycleGAN training loss curves on different settings. Note that we showed vanilla CycleGAN, Double Discriminator (3+5) CycleGAN and U-Net Discriminator CycleGAN here.}
\label{fig:train_loss}
\end{figure}

The patterns in loss curves (as shown in Figure \ref{fig:train_loss}) reveal that discriminator loss and cycle-consistency loss are decreasing by epoch. However, for all three networks, the generator loss increases after some point, which is an indicator of imbalanced training of $G$ and $D$, where $D$ is getting too strong. Notice that generator loss includes both the reconstruction loss and the adversarial part to fool discriminator, so increase in generator loss might also show that the network is learning to add more random details in order to fool discriminator. 

We can also see that there is not much difference between three different losses. This is coherent with the statement that loss is not a useful indicator of generated visual quality in GAN training.

\subsection{Discussion of Evaluation Metrics}

 It is known that GAN results are hard to evaluate as the applications are usually on the edge of art and technology. 

We looked into some evaluation metrics such as Inception score used in WGAN paper\citep{gulrajani2017improved} and FID score in the paper\citet{heusel2017gans} that both correlate well with human judgment. However, it’s important to evaluate Inception score on a large enough number of samples (i.e. 50k) and samples from different classes as part of this metric measures diversity. Alternatively, FID score captures the similarity of generated images to real ones better than the Inception Score. But it recommends using a minimum sample size of 10k to calculate the FID otherwise the true FID of the generator is underestimated. 

Since our datasets only contains one class mainly - person, and each dataset is quite small (around 2k), the inception score and FID can't tell much difference between how good the generated images are. Considering that there are not many videos to evaluate and easy to tell the difference, we basically based on manual inspection of the visual fidelity of generated videos. Link to the generated videos is given in the first section and the appendix.



\section{Conclusion and Future Work}
This project explored different approaches to generate better face-off videos based on CycleGAN\citep{zhu2017unpaired}. Our experiment shows that: using two distinct discriminators, deeper discriminator networks or applying face segmentation masks as weight to cycle-consistency loss would result in smoother and more stable face-off results.

During this project, we also found some drawbacks for existing CycleGAN structure and we want to solve the following problems in the future:
\begin{enumerate}
\item Exploring the possibility of domain transfer using CycleGAN, i.e. training transfer between different types of objects.
\item Increasing the capability of CycleGAN face-off in complicated backgrounds.
\item Extending face-off to body-off: training GANs for generating full-body pose transfer.
\end{enumerate}

\bibliography{bibliography}
\appendix

\section{Appendix}
\subsection{Russ with Great Face, Hair and Pose!}
\begin{figure}[H]
\centering
\begin{subfigure}{\textwidth}
  \centering
  \includegraphics[width=\linewidth]{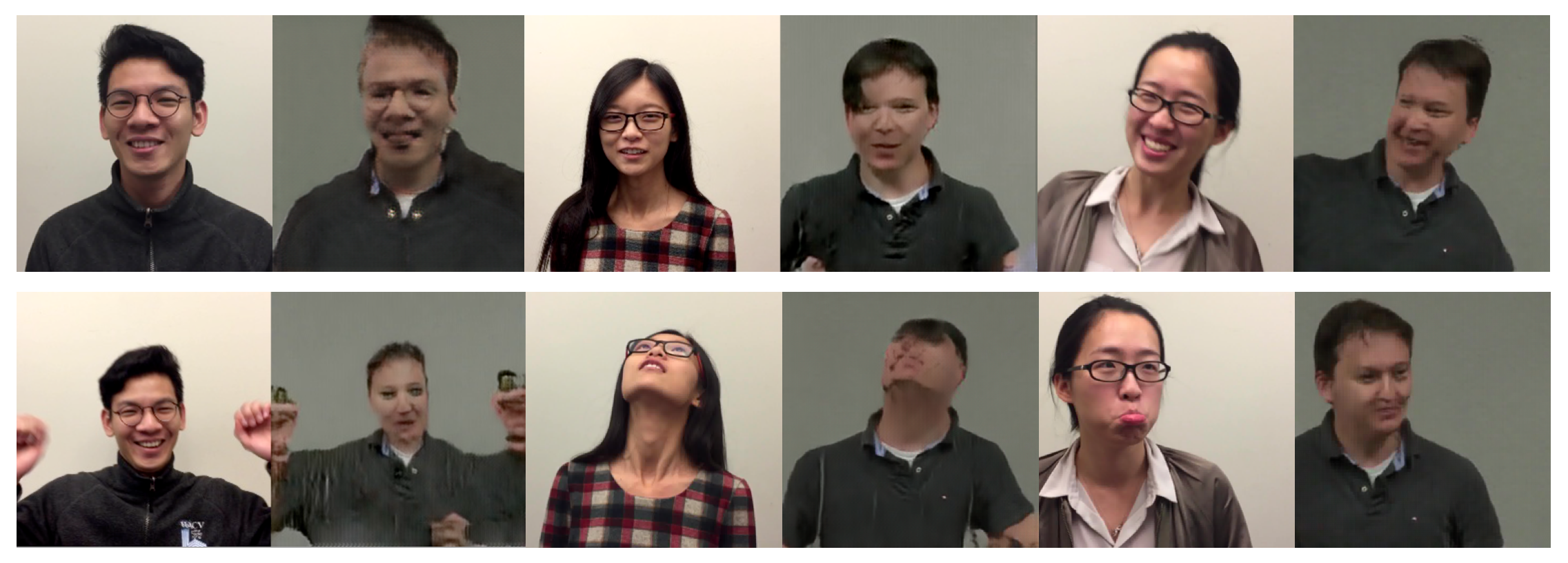}
\end{subfigure}
\label{fig:fun_results}
\end{figure}

\subsection{More Generated Videos}
Following are our test sequences and results for Shangxuan's and Xiaohan's sequence. Please use headphone!
\begin{enumerate}
\item Shangxuan's test sequence: 
\begin{enumerate}
\item \url{https://drive.google.com/file/d/1ByGBd9iPMI16GYl91piI4lDvWGCdGrof/view?usp=sharing}
\end{enumerate}
\item Xiaohan's test sequence: 
\begin{enumerate}
\item	\url{https://drive.google.com/file/d/1NIQn6uMoCAzbRnTQWQS1utJJ1ZV_156Q/view?usp=sharing}
\end{enumerate}
\end{enumerate}

\end{document}